\documentclass[journal]{IEEEtran} 

\usepackage{amsmath}

\usepackage[dvipsnames]{xcolor}

\usepackage{float}

\usepackage{graphicx} 
\graphicspath{{images/}}
\DeclareGraphicsExtensions{.pdf,.jpeg,.png}

\usepackage[caption=false]{subfig}

\usepackage{multicol}
\usepackage{multirow}

\usepackage{url}

\usepackage{comment}
\usepackage{verbatim}

\usepackage{booktabs}
\usepackage{array}

\usepackage{forest}

\usepackage{hyperref}

\usepackage{siunitx}

\usepackage[nolist]{acronym}

\usepackage{textgreek}

\begin{document}

\title{Survey of Artificial Intelligence for Card Games and Its Application to the Swiss Game Jass}
\author{
    \IEEEauthorblockN{
        \textbf{Joel~Niklaus}\IEEEauthorrefmark{1}\IEEEauthorrefmark{2}, \and
        \textbf{Michele~Alberti}\IEEEauthorrefmark{1}\IEEEauthorrefmark{2}, \and
        \thanks{\IEEEauthorrefmark{1} Both authors contributed equally to this work.}
        \textbf{Vinaychandran~Pondenkandath}\IEEEauthorrefmark{2}, \and
        \textbf{Rolf~Ingold}\IEEEauthorrefmark{2}, \and
        \textbf{Marcus~Liwicki}\IEEEauthorrefmark{2}\IEEEauthorrefmark{4}
    }\\
    \vspace{0.2cm}
    \IEEEauthorblockA{
        \IEEEauthorrefmark{2}%
        \textit{Document Image and Voice Analysis Group (DIVA)} \\
        University of Fribourg, Switzerland\\
        \{firstname\}.\{lastname\}@unifr.ch \\
        \vspace{0.15cm}
        \IEEEauthorrefmark{4}%
        \textit{Machine Learning Group} \\
        Lule{\aa} University of Technology, Sweden\\
        marcus.liwicki@ltu.se\\
    }
}

\markboth{}{First author et al. : Title}

\maketitle

\thispagestyle{empty}

\begin{acronym}[Bash]
    \acro{aka}{also known as}

    \acro{NE}{Nash Equilibrium}
    \acrodefplural{NE}{Nash Equilibria}
    
    \acro{API}{Application Programming Interface}
    
    \acro{AI}{Artificial Intelligence}
    \acro{AGI}{Artificial General Intelligence}
    \acro{RL}{Reinforcement Learning}
    
    \acro{Dota}{Defense of the Ancients}
    
    \acro{PIG}{Perfect Information Game}
    \acro{IIG}{Imperfect Information Game}
    
    \acro{MC}{Monte Carlo}
    \acro{MCTS}{Monte Carlo Tree Search}
    \acro{IS-MCTS}{Information Set Monte Carlo Tree Search}
    \acro{UCT}{Upper Confidence Bound for Trees}
    \acro{UCB}{Upper Confidence Bound}
    

    \acro{NFSP}{Neural Fictitious Self-Play}
    
    \acro{FOM}{First Order Methods}
    \acro{EGT}{Excessive Gap Technique}

    \acro{CFR}{Counterfactual Regret Minimization}
    \acro{MCCFR}{Monte Carlo Sampling for Regret Minimization}
    \acro{OOS}{Online Outcome Sampling}

    \acro{PG}{Policy Gradient}
    \acro{PPO}{Proximal Policy Optimization}
    \acro{A2C}{Advantage Actor Critic}
    
    \acro{TDL}{Temporal Difference Learning}
    
    \acro{KNN}{K-Nearest Neighbour}
    
    \acro{EA}{Evolutionary Algorithm}
    
    \acro{MLP}{Multilayer Perceptron}
    \acro{ANN}{Artificial Neural Network}
\end{acronym}

\begin{abstract}
In the last decades we have witnessed the success of applications of \acl{AI} to playing games.
In this work we address the challenging field of games with hidden information and card games in particular. 
Jass is a very popular card game in Switzerland and is closely connected with Swiss culture. 
To the  best of our  knowledge, performances of  \acl{AI}  agents  in  the  game  of  Jass  do  not outperform top players yet.
Our contribution to the community is two-fold.
First, we provide an overview of the current state-of-the-art of \acl{AI} methods for card games in general. 
Second, we discuss their application to the use-case of the Swiss card game Jass.
This paper aims to be an entry point for both seasoned researchers and new practitioners who want to join in the Jass challenge.
\end{abstract}

\section{Introduction}

The research field of \ac{AI} applied to playing games has been subject to several breakthroughs in the last years.
In particular, the branch of \acp{PIG} --- where the entire game state is known to all players at all points in time --- has seen machines triumph over human professional players in different occasions, such as for Chess, the Atari games or Go. 
When it comes to \acp{IIG} --- where part of the information is unknown to the players, such as in card games --- there is a thin line separating \ac{AI} from humans, who still have the upper hand against state-of-the-art agents.
However, recent work shows that in constrained situations, the gap between humans and \ac{AI} is becoming thinner.
This is particularly visible when considering advances on Texas hold'em no-limit poker \cite{Libratus} and the computer games \ac{Dota} 2 and StarCraft II. 

Hidden information is also present in many real world scenarios, like negotiations, surgical operations, business, physics and others.
Many of these situations can be formalized as games, which in turn can be solved using the methods refined in the test bed of card games.
Most card games involve hidden information, which makes them both a suitable and interesting domain for further research on \ac{AI}. 
There is a large variety of card games, where many use different cards and rules, which poses different challenges to the players.
To tackle these different issues, several methods have been proposed.
Unfortunately, these methods are often either very complex, or introduce only minor modifications to address a particular issue for a particular game. 
Despite producing good empirical results, this practice leads to a more complex landscape of literature which is at times hard to navigate, especially for new practitioners in the field. 
To combat this unwanted side effect, overviews of the current recent trends and methods are very helpful.
In this work, we aim to provide such an overview of \ac{AI} methods applied to card games. 
In the appendix there is a short description of the games we mention in this work. 

To complement the overview, we chose to use the card game ``Jass'' as a use-case for a discussion of the methods present above.
Jass is a very popular card game in Switzerland and tightly linked to the Swiss culture. 
From a research point of view, Jass is a challenging game because a) it is played by more than two players (specifically, four divided in two teams of two), b) it involves hidden information (the cards of the other players), c) it is difficult to master by humans and d) the number of information sets is much bigger than that of other popular card games such as Poker. 
However to the best of our knowledge, a formal approach towards Jass has not been address in a scientific manner yet.
The Swiss Intercantonal Lottery and some Jass applications have deployed some \ac{AI} agents, but these programs are not yet able to beat top human players. 
    
    \subsection*{Main Contribution}
    In this work, we aim to address a gap in the literature regarding \ac{AI} approaches towards card games with a particular emphasis on the popular Swiss card game Jass.
    To the best of our knowledge, there has not been a formal scientific approach to Jass outlined in the literature. 
    To this end, we discuss the potential merits and demerits of the different methods outlined in the paper towards Jass.

\section{Related Work}

In this section we review the relevant related work.
In their book, Yannakakis et al. \cite{GameAIBook} gave a general overview of \ac{AI} development in games, while Rubin et al. \cite{PokerReview} provided a more specific review on the methods used in computer Poker. 
In his thesis, Burch \cite{WhyIIGSAreHard} reviewed the state-of-the-art in \ac{CFR}, a family of methods very heavily used in computer Poker. 
Finally, Browne et al. \cite{SurveyMCTSMethods} surveyed the different variants of \ac{MCTS}, a family of methods used for \acp{AI} in many games of both perfect and imperfect information.
We are not aware of any work that specifically addresses the domain of card games.

\section{Theoretical Foundation}
\label{toc:theoretical_foundation}

In this section we introduce terms necessary to understand \acs{AI} for card games.

    \subsection{Game Types}
    Games can be classified in many dimensions.
    In this section we outline the ones most important for classifying card games.
    
        
    
        
            
        \subsubsection{Extensive-form Games}
        Sequential games are normally formalized as extensive-form games. 
        These games are played on a decision tree, where a node represents a decision point for a player and an edge describes a possible action leading from one state to another.
        For each node in this tree it is possible to define an information set.
            An information set includes all the states a player could be in, given the information the player has observed so far. 
            In \acp{PIG}, these information sets always only comprise exactly one state, because all information is known. 
            In an \ac{IIG} like Poker, this information set contains all the card combinations the opponents could have, given the information the player has, i.e. the cards on the table and the cards in the hand.
        
        
        
        \subsubsection{Coordination Games}
        Unlike many strategic situations, collaboration is central in a coordination game, not conflict.
        In a coordination game, the highest payoffs can only be achieved through team work. 
        Choosing the side of the road to drive on is a simple example of a coordination game.
        It does not matter which side of the road you agree on, but to avoid crashes, an agreement is essential.
        In card games, like Bridge or Jass, where there are two teams playing against each other, the interactions within the team can be seen as a coordination game.
    
    \subsection{AI Performance Evaluation}
    When developing an \ac{AI}, it is important to accurately measure its strength in comparison to other \acp{AI} and humans. The ultimate goal is to achieve optimal play.
    When a player is playing optimally, s/he does not make any mistakes but plays the best possible move in every situation. 
    When an optimal strategy in a game is known, this game is considered solved.
    
        \subsubsection{\acl{NE}}
        \label{sec:NashEquilibrium}
        A \ac{NE} describes a combination of strategies in non-cooperative games. When two or more players are playing their respective part of a \ac{NE}, any unilateral deviation from the strategy leads to a negative relative outcome for the deviating player \cite{NashEquilibrium}. 
        So when programming players for games, the goal is to get as close as possible to a \ac{NE}. 
        When one is playing a \ac{NE} strategy, the worst outcome that can happen is coming to a draw. This means that a \ac{NE} player wins against any player not playing a \ac{NE} strategy. 
        In games involving chance (the cards dealt at the beginning in the case of Poker), the player may not win every single game. 
        Thus, many games may have to be played to evaluate the strategies. 
        A \ac{NE} strategy is particularly beneficial against strong players. Therefore, it does not make any mistakes the opponent could possibly exploit. 
        On the other hand, a \ac{NE} strategy might not win over a sub-optimal player by a large margin because it does not actively try to exploit the opponent but rather tries not to commit any mistakes at all. 
        There exists a \ac{NE} for every finite game \cite{NashEquilibrium}. 
        
        \subsubsection{Exploitability}
        Exploitability is a measure for this deviation from a \ac{NE} \cite{Exploitability}. 
        The higher the exploitability, the greater the distance to a \ac{NE}, and therefore, the weaker the player.
        A \ac{NE} strategy constitutes optimal play, since there is no possible better strategy. 
        However, there are different \ac{NE} strategies which differ in their effectiveness of exploiting non-\ac{NE} strategies \cite{NERefinements}.
        If it is not possible to calculate such a strategy (for example, because the state space is too large), we want to estimate a strategy which minimizes the deviation from a \ac{NE}. 
        
        
        
        
        \subsubsection{Comparison to Humans}
        When designing \acp{AI} it is always interesting to evaluate how well they perform in comparison to humans. 
        Here we distinguish four categories: sub-human, par-human, high-human and super-human \ac{AI} which respectively mean worse than, similar to, better than most and better than all humans. 
        The current best \ac{AI} agents in Jass achieve par-human standards.
        In Bridge, current computer programs achieve expert level, which constitutes high-human proficiency.
        In many \acp{PIG} like Go or Chess, current \acp{AI} achieve super-human level.

\section{Rule-Based Systems}
Rule-based systems leverage human knowledge to build an \ac{AI} player \cite{GameAIBook}. 
Many simple \acp{AI} for card games are rule-based and then used as baseline players. 
This mostly entails a number of if-then-else statements which can be viewed as a man-made decision tree. 

Ward et al. \cite{MCSearchRuleBasedMagic} created a rule-based \ac{AI} for Magic: The Gathering which was used as a baseline player.
Robilliard et al. \cite{7WondersMCTS} developed a rule-based \ac{AI} for 7 Wonders which was used as a baseline player.
Watanabe et al. \cite{Scopone} implemented three rule-based players. The greedy player behaves like a beginner player. The other two follow more advanced strategies taken from strategy books and are behaving like expert players.
Osawa \cite{HanabiRuleBased} presented several par-human rule-based strategies for Hanabi. His results indicated that feedback-based strategies achieve higher scores than purely rational ones.
Van den Bergh et al. \cite{HanabiAspects} developed a strong par-human rule-based \ac{AI} for Hanabi.
Whitehouse et al. \cite{SpadesMCTSRuleBased} evaluated the rule-based Spades player developed by AI Factory. Based on player reviews they found it to decide weakly in certain situations but to be a strong par-human player overall.

\section{\acl{RL} Methods}

\ac{RL} is a machine learning method which is frequently used to play games. 
It consists of an agent performing actions in a given environment. 
Based on its actions, the agent receives positive rewards which reinforce desirable behaviour and negative rewards which discourage unwanted behaviour.
Using a value function, the agent tries to find out which action is the most desirable in a given state.

    \subsection{\acl{TDL}}
    \ac{TDL} updates the value function continuously after every iteration, as opposed to earlier strategies which waited until the episode's end \cite{RLIntroduction}. 
    
    Sturtevant et al. \cite{HeartsFeatureConstruction} developed a sub-human \ac{AI} for Hearts using Stochastic Linear Regression and \ac{TDL} which outperforms players based on minimax search.
    
    \subsection{\acl{PG}}
    \ac{PG} is an algorithm which directly learns a policy function mapping a state to an action \cite{RLIntroduction}. 
    \ac{PPO} is an extension to the \ac{PG} algorithm improving its stability and reducing the convergence time \cite{PPO} 

    Charlesworth \cite{Big2} applied \ac{PPO} to Big 2, reaching par-human level.
    
    \subsection{\acl{CFR}}
    \ac{CFR} \cite{CFR} is a self-playing method that works very well for \acp{IIG} and has been used by the most successful poker AIs \cite{Libratus, DeepStack}. ``Counterfactual'' denotes looking back and thinking ``had I only known then...''. 
    ``Regret'' says how much better one would have done, if one had chosen a different action.
    And ``minimization'' is used to minimize the total regret over all actions, so that the future regret is as small as possible. 
    Note that \ac{CFR} only requires memory linear to the number of information sets and not to the number of states \cite{PokerReview}. 
    Additionally, \ac{CFR} has been able to exploit non-\ac{NE} strategies computed by \ac{UCT} agents in simultaneous games \cite{UCTVsCFR}.
    
        \subsubsection{\acl{CFR}+}
        \ac{CFR}+ is a re-engineered version of \ac{CFR}, which drastically reduces convergence time.
        It always iterates over the entire tree and only allows non-negative regrets. \cite{HeadsUpLimitTexasHoldemPokerIsSolved}
        Bowling et al. \cite{HeadsUpLimitTexasHoldemPokerIsSolved} used \ac{CFR}+ to essentially solve heads-up limit Texas hold'em Poker in 2015.
        
        Morav{\v c}{\'\i}k et al. \cite{DeepStack} developed a general algorithm for imperfect information settings, called DeepStack. 
        With statistical significance, it defeated professional poker players in a study over 44000 hands.

        \subsubsection{Deep \acl{CFR}}
        Deep \ac{CFR} \cite{DeepCFR} combines \ac{CFR} with deep \acp{ANN}. 
        Brown et al. \cite{Libratus} leverage deep \ac{CFR} to decisively beat four top human poker players in 2017 with their program called Libratus. 
        
        
        \subsubsection{Discounted \acl{CFR}}
        Discounted \ac{CFR} \cite{DiscountedCFR} matches or outperforms the previous state-of-the-art variant \ac{CFR}+ depending on the application by discounting prior iterations. 
    
    \subsection{\acl{NFSP}}
    In \ac{NFSP}, two players start with random strategies encoded in an \ac{ANN}. 
    They play against each other knowing the other player's strategy improving the own strategy. 
    With an increasing number of iterations, the strategies typically approach a \ac{NE}.
    Since \ac{NFSP} \cite{NFSP} has a slower convergence rate than \ac{CFR} it is not widely used.
    
    Heinrich et al. \cite{NFSP} applied \ac{NFSP} to Texas hold'em Poker and reported similar performance to the state-of-the-art super-human programs.
    In Leduc Poker, a simplification of the former, they approached a \ac{NE}.
    Kawamura et al. \cite{NFSPMultiplayerImperfectInformationGames} calculated approximate \ac{NE} strategies with \ac{NFSP} in multiplayer \acp{IIG}.
    
    \subsection{\acl{FOM}}
    \ac{FOM} like \ac{EGT} are, like \ac{CFR}, methods which approximate \ac{NE} strategies in \acp{IIG}. 
    They have a better theoretical convergence rate than \ac{CFR} because of lower computational and memory costs.
    Note that, like \ac{CFR}, \ac{EGT} is only able to approach a \ac{NE} in two-player games \cite{ExcessiveGapTechnique}.
    
    Kroer et al. \cite{ExcessiveGapTechnique} applied a variant of \ac{EGT} to Poker reporting faster convergence than some \ac{CFR} variants.
    They argue that, given more hyper parameter tuning, the performance of \ac{CFR}+ can be reached.

\section{\acl{MC} Methods}

\ac{MC} methods use randomness to solve problems that might be deterministic in principle.

    \subsection{\acl{MC} Simulation}
    \label{sec:MCSimulation}
    \ac{MC} Simulation uses a large number of random experiments to numerically solve large problems involving many random variables.
    
    Mitsukami et al. \cite{Mahjong} developed a par-human \ac{AI} for Japanese Mahjong using \ac{MC} Simulation.
    Kupferschmid et al. \cite{SkatMonteCarloSimulation} applied \ac{MC} Simulation to Skat to obtain the game-theoretical value of a Skat hand. 
    Note that they converted the game to a \ac{PIG} by making all the cards known.
    Yan et al. \cite{SolitaireRuleBased} report a 70\% win rate using \ac{MC} Simulation in a Klondike version, which has all cards revealed to the player. Note that this converts the game to a \ac{PIG}.
    
    \subsection{Flat \acl{MC}}
    Flat \ac{MC} uses \ac{MC} Simulation, with the actions in a given state being uniformly sampled \cite{SurveyMCTSMethods}. 
    
    Ginsberg \cite{BridgeFlatMC} achieves world champion level play in Bridge using Flat \ac{MC} in 2001.

    \subsection{\acl{MCTS}}

    \ac{MCTS} consists of four stages: Selection, Expansion, Simulation and Backpropagation \cite{SurveyMCTSMethods}. 
    Selection: Starting from the root node, an expandable child node is selected. 
    A node is expandable if it is non-terminal (i.e. it does have children) and has unvisited children. \\
    Expansion: The tree is expanded by adding one or more child nodes to the previously selected node. \\
    Simulation: From these new children nodes a simulation is run to acquire a reward at a terminal node. \\
    Backpropagation: The simulation's result is used to update the information in the affected nodes (nodes in the selection path).
    A tree policy is used for selecting and expanding a node and the simulation is run according to the default policy.
    
    Browne et al. \cite{SurveyMCTSMethods} gives a detailed overview of the \ac{MCTS} family . 
    In this section we outline the variants used on card games.
        
        \subsubsection{\acl{UCT}}
        \ac{UCT} is the most common \ac{MCTS} method, using upper confidence bounds as a tree policy, which is a formula that tries to balance the exploration/exploitation problem \cite{UCT}. 
        When the search explores too much, the optimal moves are not played frequently enough and therefore it may find a sub-optimal move. 
        When the search exploits too much, it may not find a path which promises much greater payoffs and it therefore also may find a sub-optimal move. 
        Minimax is a basic algorithm used for two-player zero-sum games, operating on the game tree. When the entire tree is visited, minimax is optimal \cite{GameAIBook}. 
        \ac{UCT} converges to minimax given enough time and memory \cite{UCT}.
        
        Sievers et al. \cite{Doppelkopf} applied \ac{UCT} to Doppelkopf reaching par-human performance.
        Sch{\"a}fer \cite{SkatUCT} used \ac{UCT} to build an \ac{AI} for Skat, which is still sub-human but comparable to the \ac{MC} Simulation based player proposed by Kupferschmid et al. \cite{SkatMonteCarloSimulation}.
        Swiechowski et al. \cite{HearthStoneCombination} combined an \ac{MCTS} player with supervised learning on the logs of sample games, achieving par-human performance. 
        Santos et al. \cite{MCTSExperimentsHearthstone} outperformed basic \ac{MCTS} based \acp{AI} by combining it with domain-specific knowledge.
        Heinrich et al. \cite{SmoothUCT} combined \ac{UCT} with self-play and apply it to Poker. 
        They reported convergence to a \ac{NE} in a small Poker game and argue that, given enough training, convergence can also be reached in large limit Texas Hold'em Poker.
        
        \subsubsection{Determinization}
        Determinization 
        is a technique which allows solving an \ac{IIG} with methods used for \acp{PIG}. 
        Determinization samples many states from the information set and plays the game to a terminal state based on these states of perfect information.
        
        Bjarnason et al. \cite{Klondike} studied Klondike using \ac{UCT}, hindsight optimization and sparse sampling. 
        Hindsight optimization uses determinization and hindsight knowledge to improve the strategy. 
        They developed a policy which wins at least 35\% of games, which is a lower bound for an optimal Klondike policy.
        Sturtevant \cite{UCTMultiplayerGames} applied \ac{UCT} with determinization to the multiplayer games Spades and Hearts. He reported similar performance to the state-of-the-art at that time in Spades and slightly better performance in Hearts.
        Cowling et al. \cite{Magic} applied \ac{MCTS} with determinization approaches to the card game Magic: The Gathering achieving high-human performance and outperforming an expert-level rule-based player.
        Robilliard et al. \cite{7WondersMCTS} applied \ac{UCT} with determinization to 7 Wonders outperforming rule-based \acp{AI}. 
        The experiments against human players were promising but not statistically significant. 
        Solinas et al. \cite{SkatSupervised} used \ac{UCT} and supervised learning to infer the cards of the other players, improving over the state-of-the-art in Skat card-play.
        Edelkamp \cite{ChallengingHumanSupremacySkat} combined distilled expert rules, winning probabilities aggregations and a fast tree exploration into an \ac{AI} for the Mis\`ere variant of Skat significantly outperforming human experts.

        \subsubsection{\acl{IS-MCTS}}
        \ac{IS-MCTS} tackles the problem of strategy fusion which includes the false assumption that different moves can be taken from different states in the information set \cite{IS-MCTS}.
        However, because the player does not know of the different states in the information set, it cannot decide differently, based on different states.
        \ac{IS-MCTS} operates directly on a tree of information sets.
    
        Whitehouse et al. \cite{DouDiZhuDeterminization} used \ac{MCTS} with determinization and information sets on Dou Di Zhu. 
        They did not report any significant differences in performance between the two proposed algorithms.
        Watanabe et al. \cite{Scopone} presented a high-human \ac{AI} using \ac{IS-MCTS} for the Italian card game Scopone which consistently beat strong rule-based players.
        Walton-Rivers et al. \cite{Hanabi} applied \ac{IS-MCTS} to Hanabi, but they measured inferior performance to rule-based players.
        Whitehouse et al. \cite{SpadesMCTSRuleBased} found an \ac{MCTS} player to be stronger than rule-based players in the card game Spades.
        They integrated \ac{IS-MCTS} with knowledge-based methods to create more engaging play. 
        Cowling et al. \cite{SpadesAnalysis} performed a statistical analysis over 27592 played games on a mobile platform to evaluate the player's difficulty for humans.
        Devlin et al. \cite{SpadesMCTSGameplayData} combined insights from game play data with \ac{IS-MCTS} to emulate human play.

    \subsection{\acl{MCCFR}}
    \ac{MCCFR} drastically reduces the convergence time of \ac{CFR} by using \ac{MC} Sampling \cite{MCCFR}. 
    \ac{MCCFR} samples blocks of paths from the root to a terminal node and then computes the immediate counterfactual regrets over these blocks.
    
    Lanctot et al. \cite{MCCFR} showed this faster convergence rate in experiments on Goofspiel and One-Card-Poker. 
    Ponsen et al. \cite{CFRvsMCTS} evidences that \ac{MCCFR} approaches a \ac{NE} in Poker. 
    
    \subsubsection{\acl{OOS}}
    \ac{OOS} is an online variant to \ac{MCCFR} which can decrease its exploitability with increasing search time \cite{OOS}.
     
    Lis\'{y} et al. \cite{OOS} demonstrated that \ac{OOS} can exploit \ac{IS-MCTS} in Poker knowing the opponent's strategy and given enough computation time.

\section{\aclp{EA}}

\acp{EA} are inspired by evolutionary theory. Strong individuals --- strategies in the case of game \acp{AI} --- can survive and reproduce, whereas weaker ones eventually become extinct \cite{GameAIBook}.

Mahlmann et al. \cite{Dominion} compared three \ac{EA} agents with different fitness functions in Dominion. 
They argued that their method can be used for automatic game design and game balancing.
Noble \cite{PokerGA} applied a \ac{EA} evolving \acp{ANN} to Poker in 2002 improving over the state-of-the-art at the time.

\section{Use-Case: Swiss Card Game Jass}

Jass is a trick-taking traditional Swiss card game often played at social events. 
It involves hidden information, is sequential, non-cooperative, finite and constant-sum, as there are always 157 points possible in each game. 
The Swiss Intercantonal Lottery provide a guide for general Jass rules\footnote{\url{www.swisslos.ch/en/jass/informations/jass-rules/principles-of-jass.html}} and for the variant Schieber in particular\footnote{\url{www.swisslos.ch/en/jass/informations/jass-rules/schieber-jass.html}}.

    
    \subsection{Coordination Game Within Jass}
    Schieber is a non-cooperative game, since the two teams are opposing each other.
    However, additionally, the activity within a team can be formulated as a coordination game.
    This adds another dimension to the game as it enables cooperation between the players within the game to maximize the team's benefit. 
    Although the rules of the game forbid any communication during a game within the team, by playing specific cards in certain situations, the two players can convey information about the cards they have. 
    For this to work, of course they must have the same understanding of this communication by card play. 
    Humans have some existing ``agreements'' like a ``discarding policy''.
    Discarding tells the partner which suits the player is bad at. 
    It is interesting to investigate, whether \acp{AI} are able to pick up these ``agreements'' or even come up with new ones.

    \subsection{Suitable Methods for \acp{AI} in Card Games by the Example of Jass}

    \ac{MCTS} and \ac{CFR} are the two families of algorithms that have most successfully been applied to card games. 
    In this section we are comparing these two methods' advantages and disadvantages in detail by the example of the trick-taking card game Jass.

    To the best of our knowledge, \ac{CFR} has almost exclusively been applied to Poker so far, although the authors claim that it can be applied to any \ac{IIG} \cite{DeepCFR}. 
    \ac{CFR} provides theoretical guarantees for approaching a \ac{NE} in two player \acp{IIG} \cite{CFR}.
    On the other hand, as we discussed in section \ref{sec:NashEquilibrium}, pure \ac{NE} strategies may not be able to specifically exploit weak opponents. 
    Additionally, \ac{CFR} needs a lot of time to converge, compared to \ac{MCTS} \cite{CFRvsMCTS}.

    \ac{MCTS} has been applied to a plethora of complex card games including Bridge, Skat, Doppelkopf or Spades, as we have illustrated in the previous sections. 
    It finds good strategies fast but only converges to a \ac{NE} in \acp{PIG} and not necessarily in \acp{IIG} \cite{CFRvsMCTS}. 
    As opposed to \ac{CFR}, \ac{MCTS} does not find the moves with lowest exploitability, but the ones with highest chance of winning \cite{ConvergenceMCTS}. 
    \ac{MCTS} eventually converges to minimax, but total convergence is infeasible for large problems \cite{SurveyMCTSMethods}.
    
    So, if the goal is to find a good strategy relatively fast, \ac{MCTS} should be chosen, whereas \ac{CFR} should be selected, if the goal is to be minimally exploitable \cite{CFRvsMCTS}. 
    To put it simply, \ac{CFR} is great at not losing, but not very good at destroying an opponent and \ac{MCTS} is great at finding good strategies fast, but not very good at resisting against very strong opponents.


    \subsection{Preliminary Results}
    Preliminary experiments not presented in this paper show that \ac{MCTS} is a promising approach for a strong \ac{AI} playing Jass.


    

\section{Conclusion}
In this paper we first provided an overview of the methods used in \ac{AI} development for card games. 
Then we discussed the advantages and disadvantages of the two most promising families of algorithms (\ac{MCTS} and \ac{CFR}) in more detail. 
Finally, we presented an analysis for how to apply these methods to the Swiss card game Jass.

\bibliography{biblio}
\bibliographystyle{IEEEtran}

\appendix
\section{Game Descriptions}
In this section we give the gist of the less well-known games discussed in the paper (in order of appearance).

\emph{Magic: The Gathering} is a trading and digital collectible card game played by two or more players.
\emph{7 Wonders} is a board game with strong elements of card games including hidden information for two to seven players.
\emph{Scopone} is a variant of the Italian card game Scopa. 
\emph{Hanabi} is a French cooperative card game for two to five players.
\emph{Spades} is a four player trick-taking card game mainly played in North America.
\emph{Big 2} is a Chinese card game for two to four players mainly played in East and South East Asia. 
The goal is, to get rid of all of one's cards first. 
\emph{Mahjong} is a traditional Chinese tile-based game for four (or seldom three) players similar to the Western game Rummy.
\emph{Skat} is a three player trick-taking card game mainly played in Germany.
\emph{Klondike} is a single-player variant of the French card game Patience and shipped with Windows since version 3.
\emph{Bridge} is a trick-taking card game for four players played world-wide in clubs, tournaments, online and socially at home.
It has often been used as a test bed for \ac{AI} research and is still an active area of research, since super-human performance has not been achieved yet.
\emph{Doppelkopf} is a trick-taking card game for four people, mainly played in Germany.
\emph{Hearthstone} is an online collectible card video game, developed by Blizzard Entertainment. 
\emph{Hearts} is a four player trick-taking card game, mainly played in North America.
\emph{Dou Di Zhu} is a Chinese card game for three players.
\emph{Goofspiel} is a simple bidding card game for two or more players.
\emph{One-Card-Poker} generalizes the minimal variant Kuhn-Poker.
\emph{Dominion} is a modern deck-building card game similar to Magic: The Gathering.

\end{document}